\documentclass[sigconf]{aamas}  

\usepackage{amsmath}
\usepackage{amsfonts}
\usepackage{amssymb}
\usepackage{subcaption}
\usepackage{color}
\usepackage{algorithm}
\usepackage{algpseudocode}
\usepackage{changes}
\definechangesauthor[color=blue]{ST}
\definechangesauthor[color=olive]{LSW}
\definechangesauthor[color=purple]{ng}
\definechangesauthor[color=orange]{Reviewer1}
\definechangesauthor[color=teal]{Reviewer2}
\newif\iffinal
\finalfalse

\iffinal
\usepackage[disable]{todonotes}
\newcommand{\note}[2][]{}
\newcommand{\stnote}[1]{}
\newcommand{\lswnote}[1]{}
\newcommand{\ngnote}[1]{}
\newcommand{\revonenote}[1]{}
\newcommand{\revtwonote}[1]{}
\else
\usepackage[colorinlistoftodos]{todonotes}
\makeatletter
\setremarkmarkup{\todo[color=Changes@Color#1!20,size=\scriptsize]{#1: #2}}
\makeatother
\newcommand{\note}[2][]{\added[#1,remark={#2}]{}}
\newcommand{\stnote}[1]{\note[id=ST]{#1}}
\newcommand{\lswnote}[1]{\note[id=LSW]{#1}}
\newcommand{\ngnote}[1]{\note[id=ng]{#1}}
\newcommand{\revonenote}[1]{\note[id=Reviewer1]{#1}}
\newcommand{\revtwonote}[1]{\note[id=Reviewer2]{#1}}
\fi

\DeclareMathOperator*{\argmax}{arg\,max}

\setcopyright{ifaamas}  
\acmDOI{}  
\acmISBN{}  
\acmConference[AAMAS'18]{Proc.\@ of the 17th International Conference on Autonomous Agents and Multiagent Systems (AAMAS 2018)}{July 10--15, 2018}{Stockholm, Sweden}{M.~Dastani, G.~Sukthankar, E.~Andr\'{e}, S.~Koenig (eds.)}  
\acmYear{2018}  
\copyrightyear{2018}  
\acmPrice{}  

\begin{document}

\title{Deep Abstract Q-Networks}  



\author{Melrose Roderick}
\affiliation{%
  \institution{Carnegie Mellon University}
  \city{Pittsburgh}
  \state{Pennsylvania} 
  \postcode{15213}
}
\email{mroderick@cmu.edu}

\author{Christopher Grimm}
\affiliation{%
  \institution{University of Michigan}
  \city{Ann Arbor} 
  \state{Michigan} 
  \postcode{48109}
}
\email{crgrimm@umich.edu}

\author{Stefanie Tellex}
\affiliation{%
  \institution{Brown University}
  \city{Providence} 
  \state{Rhode Island} 
  \postcode{02912}
}
\email{stefie10@cs.brown.edu}

\begin{abstract}  
We examine the problem of learning and planning on high-dimensional domains with long horizons and sparse rewards. 
Recent approaches have shown great successes in many Atari 2600 domains.
However, domains with long horizons and sparse rewards, such as Montezuma's Revenge and Venture, remain challenging for existing methods.
Methods using abstraction \citep{dietterich2000hierarchical,sutton1999between} have shown to be useful in tackling long-horizon problems.
We combine recent techniques of deep reinforcement learning with existing model-based approaches using an expert-provided state abstraction.
We construct toy domains that elucidate the problem of long horizons, sparse rewards and high-dimensional inputs, and show that our algorithm significantly outperforms previous methods on these domains.
Our abstraction-based approach outperforms Deep Q-Networks \citep{mnih2015human} on Montezuma's Revenge and Venture, and exhibits backtracking behavior that is absent from previous methods.
\end{abstract}

%


\maketitle


\section{Introduction}
Recent advances in deep learning have enabled the training of reinforcement learning  agents in high-dimensional domains. This was most popularly demonstrated by \citet{mnih2015human} in their research into training Deep Q-Networks to play various Atari 2600 games.
While the performance attained by \citeauthor{mnih2015human} spans an impressive subset of the Atari 2600 library, several complicated games remain out of reach from existing techniques, including the notoriously difficult Montezuma's Revenge (MR) and Venture.
These anomalously difficult domains exhibit sparse reward signals and sprawling partially-observable mazes.
The confluence of these traits produces difficult games beyond the capabilities of existing deep techniques to solve.
In spite of these considerable challenges, these games are some of the closest analogs to real-world robotics problems since they require an agent to navigate a complex, unknown environment and manipulate objects to achieve long-term goals.

As an example of a long-horizon problem, consider a domain in which an agent is tasked with navigating through a series of cluttered rooms with only visual input.
The door to enter the desired room is locked and the key is at a known location in another room in this domain. 
The agent must navigate through several rooms to find the key before retracing its steps to the door to unlock it.
Learning to navigate each individual room is on its own challenging, but learning a policy to traverse multiple such rooms is much harder.

While a complete solution is presently out of reach, there have been a number of promising attempts at improving the long-term planning of deep reinforcement learning agents.
These approaches can be divided into two categories: 
\begin{enumerate}

\item Those that intrinsically motivate an agent to explore portions of the state-space that exhibit some form of novelty \citep{Bellemare2016UnifyingCE}.
\item Those that exploit some kind of abstraction to divide the learning problem into more manageable subparts \citep{Kulkarni2016HierarchicalDR,Vezhnevets2017FeUdalNF}.
\end{enumerate}

Both of these approaches suffer drawbacks. 
Novelty-based approaches indeed encourage exploration. However, this intrinsic drive toward under-explored states tends to interfere with an agent's ability to form long-term plans.
As a result, the agent may be able to find the key in the rooms but is unable to make a plan to pick up the key and then use it to unlock the door.

Abstraction-based approaches focus on end-to-end learning of both the abstractions and the resulting sub-policies, and are hindered by an extremely difficult optimization problem that balances constructing a good abstraction while still exploring the state-space and learning the policies to navigate the abstraction while the abstraction continues to change.
Moreover, given the lack of strong theoretical underpinnings for the ``goodness'' of an abstraction, little external guidance can be provided for any such optimization scheme.

To tackle domains with long horizons and sparse rewards, we propose the following method in which an experimenter provides a lightweight abstraction consisting of factored high-level states to the agent.
We then employ the formalism of the Abstract Markov Decision Process (AMDP) \citep{gopalan17} to divide a given domain into a symbolic, high-level representation for learning long-term policies and a pixel-based low-level representation to leverage the recent successes of deep-learning techniques.
In our toy example, the high-level representation would be the current room of the agent and whether the agent has the key, and the low-level representation would be the pixel values of the image.
The aforementioned factoring decomposes this symbolic, high-level state into collections of \textit{state-attributes} with associated \textit{predicate functions} in a manner similar to Object Oriented MDPs \citep{diuk2008object}.
This factoring allows us to treat actions in our high-level domain as changes in attributes and predicates rather than as state-to-state transitions, while avoiding a combinatorial explosion in the action space as the number of objects increases.
For example, once a key is retrieved, the agent should not have to re-learn how to navigate from room to room; holding a key should not generally change the way the agent navigates.

In this work, we detail our method for combining recent techniques of deep reinforcement learning with existing model-based approaches using an expert-provided state abstraction.
We then illustrate the advantages of this method on toy versions of the room navigation task, which are designed to exhibit long horizons, sparse reward signals, and high-dimensional inputs.
We show experimentally that our method outperforms Deep Q-Networks (DQN) and competing novelty-based techniques on these domains.
Finally, we apply our approach to Atari 2600 \citep{bellemare13arcade} Montezuma's Revenge (MR) and Venture and show it outperforms DQN and exhibits backtracking behavior that is absent from previous methods.

\section{Related Work}

We now survey existing long-horizon learning approaches including abstraction, options, and intrinsic motivation. 

Subgoals and abstraction are common approaches for decreasing problem horizons, allowing agents to more efficiently learn and plan on long-horizon domains.
One of the earliest reinforcement learning  methods using these ideas is MAXQ \citep{dietterich2000hierarchical}, which decomposes a \textit{flat} MDP into a hierarchy of \textit{subtasks}.
Each subtask is accompanied by a \textit{subgoal} to be completed.
The policy for these individual subtasks is easier to compute than the entire task.
Additionally, MAXQ constrains the choice of subtasks depending on the context or parent task.
A key drawback to this method is that the plans are computed recursively, meaning the high-level learning algorithm must recur down into the subtrees at training time.
This limitation forces the use of a single learning algorithm for both the high-level and low-level.
Our approach avoids this problem, allowing us to use deep reinforcement learning algorithms on the low-level to handle the high-dimensional input and model-based algorithms on the high-level to create long-term plans and guide exploration.

Temporally extended actions \citep{mcgovern1997roles} and options \citep{sutton1999between} are other commonly used approaches to decreasing problem horizons, which bundles reusable segments of plans into single actions that can be used alongside the environment actions.
Learning these options for high-dimensional domains, such as Atari games, is challenging and has only recently been performed by Option-Critic \citep{bacon2017option}.
Option-Critic, however, fails to show improvements in long-horizon domains, such as Montezuma's Revenge and Venture.
In our work we seek to learn both the sub-policies and the high-level policy.

Some existing approaches have sought to learn both the options and high-level policies in parallel.
The hierarchical-DQN (h-DQN) \citep{Kulkarni2016HierarchicalDR} is a two-tiered agent using Deep Q-Learning.
The h-DQN is divided into a low-level \textit{controller} and a high-level \textit{meta-controller}.
It is important to note that these tiers operate on different timescales, with the meta-controller specifying long-term, manually-annotated goals for the controller to focus on completing in the short-term.
These manually-annotated goals are similar to the abstraction we provide to our agent: the goals in our case would be adjacent high-level states.
However, although this method does perform action abstraction, it does not perform state abstraction.
Thus, the high-level learner still must learn over a massive high-dimensional state-space.
Our approach, on the other hand, takes advantage of both state and action abstraction, which greatly decreases the high-level state-space allowing us to use a model-based planner at the high-level.
This pattern of a high-level entity providing goal-based rewards to a low-level agent is also explored in \citet{Vezhnevets2017FeUdalNF} with the FeUdal Network.
Unlike the h-DQN, the FeUdal Network does not rely on experimenter-provided goals, opting to learn a low-level \textit{Worker} and a high-level \textit{Manager} in parallel, with the Manager supplying a vector from a learned goal-embedding to the worker.
While this method was able to achieve a higher score on Montezuma's Revenge than previous methods, it fails to explore as many rooms as novelty-based methods.
In contrast, our approach provides the abstraction to the agent, allowing us to leverage existing model-based exploration algorithms, such as R-Max \citep{brafman2002r}, which enable our agent to create long-term plans to explore new rooms.

In addition to methods that rely on a goal-based form of reward augmentation, there has been work on generally motivating agents to explore their environment.
Particularly, \citet{Bellemare2016UnifyingCE} derive a \textit{pseudo-count} formula which approximates naively counting the number of times a state occurs.
These pseudo-counts generalize well to high-dimensional spaces and illuminate the degree to which different states have been explored.
Using this information, \citet{Bellemare2016UnifyingCE} are able to produce a reward-bonus to encourage learning agents to visit underexplored states; this method is referred to as Intrinsic Motivation (IM).
This approach is shown to explore large portions of MR (15/24 rooms).
While this method is able to explore significantly better than DQN, it still fails to execute plans that required to complete MR, such as collecting keys to unlock doors.

For example, in MR, after collecting its first key, the agent ends its current life rather than retracing its steps and unlocking the door, allowing it to retain the key while returning to the starting location, much closer to the doors.
This counterintuitive behavior occurs because the factorization of the state-space in \citet{Bellemare2016UnifyingCE} renders the presence of the key and the agent's position independent, resulting in the pseudo-counts along the path back to the door still being relatively large when compared to states near the key.
Thus, the corresponding exploration bonuses for backtracking are lower than those for remaining near the key.
Therefore, if the environment terminated after a single life, this method would never learn to leave the first room. 
This  phenomenon is illustrated in our single-life MR results in Figure \ref{fig:reward_results}.
Similarly, in Venture once the IM agent has collected an item from one of the rooms, the novelty of that room encourages it to remain in that room instead of collecting all four items and thereby completing the level.
In contrast, our method allows the agent to learn a different policy before it collects the key or item and after, in order to systematically find the key or item and explore farther without dying.

Schema Networks \citep{kansky2017schema} used a model-based, object-oriented approach to improve knowledge transfer across similar Atari domains, requiring much less experience to perform well in the novel domains.
This method, however, is not able to learn from high-dimensional image data and provides no evidence of improving performance on long-horizon domains.

\section{Framework and Notation}
The domains considered in this work are assumed to be Markov Decision Processes (MDPs), defined as the tuple:
\begin{equation} \label{eq:MDP}
\langle \mathcal S, \mathcal A, \mathcal R, \mathcal T, \mathcal E \rangle 
\end{equation}
where $\mathcal S$ is a set of states, $\mathcal{A}$ is a set of actions that can be taken, ${\mathcal R} (s, a, s^\prime)$ is a function representing the reward incurred from transitioning from state $s$ to state $s^\prime$ by taking action $a$, ${\mathcal T} (s, a, s^\prime)$ is a function representing the probability of transitioning from $s$ to $s^\prime$ by taking action $a$, and ${\mathcal E} \subset {\mathcal S}$ is a set of terminal states that, once reached, prevent any future action.
Under this formalism, an MDP represents an environment which is acted upon by an agent. 
The agent takes actions from the set $\mathcal{A}$ and receives a reward and an updated state from the environment. 
In reinforcement-learning problems, agents aim to learn policies, $\pi(s) : \mathcal{S} \to \mathcal{A}$ ,  to maximize their reward over time. Their success at this is typically measured as the \textit{discounted reward} or \textit{value} of acting under a policy from a given state:
\begin{equation}
V(s) = \mathbb{E}\left[ r_t + \gamma r_{t+1} + \gamma^2 r_{t+2} + \cdots \right | \pi]
\end{equation}
where the $(r_t)$ is a sequence of random variables representing the reward of an agent acting under policy $\pi$ over time, and $\gamma \in (0, 1)$ is a discount factor applied to future reward-signals.

To allow our agent to learn and plan on an abstract level, we employ the Abstract Markov Decision Process (AMDP) formalism presented in \citet{gopalan17}.
An AMDP is a hierarchy of MDPs allowing for planning over environments at various levels of abstraction. Formally, a node in this hierarchy is defined as an augmented MDP tuple:
\begin{equation*}
\langle \tilde{{\mathcal S}}, \tilde{{\mathcal A}}, \tilde{{\mathcal T}}, \tilde{{\mathcal R}}, \tilde{{\mathcal E}}, F \rangle.
\end{equation*}
where $\tilde{\mathcal{S}}$, $\tilde{\mathcal{A}}$, $\tilde{\mathcal{T}}$, $\tilde{\mathcal{R}}$ and $\tilde{\mathcal{E}}$ mirror the standard MDP components defined in Eq. \ref{eq:MDP}, $F : {\mathcal S} \rightarrow \tilde{{\mathcal S}}$ is a {\em state projection} function that maps lower-level states in $\mathcal{S}$ to their abstract representations one-level above in the hierarchy, $\tilde{\mathcal{S}}$, and every $\tilde{a} \in \tilde{\mathcal{A}}$ represents another augmented MDP or a base environment action.

As a concrete example, consider an environment containing four connected rooms.
A simple two-tiered AMDP hierarchy might treat entire rooms as abstract states that can be transitioned between. 
Each action at the high-level would be a low-level MDP with the goal of transitioning from one room to the next. 
The action-set for these MDPs would be environment-level actions (such as UP, DOWN, LEFT, RIGHT) and a reward function would be $1$ for a successful transition and a $0$ otherwise.

\section{Model}
We now describe our hierarchical system for learning agents that exhibit long-term plans. Our approach involves learning two coupled agents simultaneously: a high-level $L_1$-agent and a low-level $L_0$-agent.
The AMDP framework allows for more levels of abstraction, but we think $2$ levels of abstraction is sufficient for our domains.

The $L_0$-agent operates on states received directly from the environment and the $L_1$-agent operates on an abstraction provided by the experimenter.
This abstraction is intended to be coarse, meaning that only limited information about the environment is provided to the $L_1$-agent and many environment states cluster into a single $L_1$ state.
The coarseness of the abstraction allows for minimal engineering on the part of the experimenter.
We use the AMDP formalism described above, defining the $L_1$-agent's environment as the MDP, $\langle \tilde{\mathcal{S}}, \tilde{\mathcal{A}}, \tilde{\mathcal{T}}, \tilde{\mathcal{R}}, \tilde{\mathcal{E}} \rangle$, and the $L_0$-agent's environment as the MDP, $\langle \mathcal{S}, \mathcal{A}, \mathcal{T}, \mathcal{R}, \mathcal{E}\rangle$.
We also denote the state projection function mapping $L_0$-states to corresponding $L_1$-states as $F: \mathcal{S} \mapsto \tilde{\mathcal{S}}$. 

\subsection{Abstract States and Actions}

To allow our agent to plan at a higher level, we project the ground level states (e.g. Atari frames) into a much lower-dimensional \textit{abstraction} for the $L_1$-agent.
Similar to Object Oriented MDPs \citep{diuk2008object}, the $L_1$-agent's abstraction is specified by:
a set of abstract states factored into \textit{attributes} that represent independent state components and
a set of \textit{predicate functions} that are used to specify dependencies or interactions between particular values of the attributes.
This information is provided to the agent in the form of a state projection function, $F: \mathcal{S} \mapsto \tilde{\mathcal{S}}$, which grounds abstract states to sets of environment states.
More precisely, let $N \in \mathbb{Z}^+$ be the number of attributes in each abstract state, $M \in \mathbb{Z}^+$ be the number of predicate functions and $\tilde{\mathcal{S}}$ be the set of provided abstract states.
For any $\tilde{s} \in \tilde{\mathcal{S}}$  we will alternatively write $(\tilde{s}_1, \ldots, \tilde{s}_N)$, to emphasize the $N$ factors of $s$.
We write $(p_1, \ldots, p_M)$ to denote the $M$ predicate functions, where each $p_j : \tilde{\mathcal{S}} \mapsto \{0, 1\}$ for $j \in 1, \ldots, M$. 
For example, the $L_1$ state space for MR (an Atari navigation task with rooms, doors, and keys) would consist of the attributes $\langle$\texttt{Agent loc}$\rangle$, $\langle$\texttt{Num keys}$\rangle$, $\langle$\texttt{i'th Key collected}$\rangle$, $\langle$\texttt{j'th Door unlocked}$\rangle$ and predicates $\langle$\texttt{Near uncollected i'th Key}$\rangle$, $\langle$\texttt{Near unlocked j'th Door}$\rangle$, $\langle$\texttt{Near locked j'th Door with key}$\rangle$ for all $i$ and $j$.

This factorization prevents our state-action space from growing combinatorially in the number of objects.
In an unfactored domain, an action that is taken with the intent of transitioning from state $S_1$ to state $S_2$ can be thought of symbolically as the ordered pair: $(S_1, S_2)$.
Since there is no predefined structure to $S_1$ or $S_2$, any variation in either state, however slight, mandates a new symbolic action.
This is particularly expensive for agents acting across multiple levels of abstraction that need to explicitly learn how to perform each symbolic action on the low-level domain.
We mitigate this learning-cost through the factorization imposed by our abstraction-attributes.
For a given state $(\tilde{s}_1, \ldots, \tilde{s}_M) \in \tilde{\mathcal{S}}$, if we assume that each $s_i$ is independent then we can represent each $L_1$-action $\tilde{a} \in \tilde{\mathcal{A}}$ as a the ordered set of intended attribute changes by performing $a$. We refer to this representation as an \textit{attribute difference} and define it formally as a tuple with $M$ entries:

\begin{equation}
\texttt{Diff}(\tilde{s}, \tilde{s}^\prime)_i \triangleq \begin{cases}
(\tilde{s}_i, \tilde{s}^\prime_i) &\text{ if } \tilde{s}_i \neq \tilde{s}^\prime_i \\
\emptyset &\text{ else}.
\end{cases}
\end{equation}
In practice, it is seldom the case that each of the abstract attributes is completely independent. To allow for modeling dependencies between certain attributes, we use the predicate functions described above and augment our previous notion of $L_1$-actions with independent attributes, representing actions as  tuples of attribute differences and evaluated predicate functions: $(\texttt{Diff}(s, s^\prime)$, $p_1(s)$, $\ldots$, $p_L(s))$ $\in$ $\tilde{\mathcal{A}}$.
In our example from above, this allows the agent to have different transition dynamics for when the doors in the room are open or closed or when the key in the room has been collected or not.
For rooms with no doors or keys, however, the transition dynamics remain constant for any configuration of unlocked doors and collected keys in the state.

\subsection{Interactions Between $L_1$ and $L_0$ Agents}
In order for the $L_0$ agents to learn to transition between $L_1$ abstract states, we need to define the $L_0$ reward function in terms of $L_1$ abstract states.
It is important to note that, much like in \citet{Kulkarni2016HierarchicalDR}, the $L_1$-agent operates at a different temporal scale than the $L_0$-agent.
However, unlike \citet{Kulkarni2016HierarchicalDR}, the $L_0$ and $L_1$-agents operate on different state-spaces, so we need to define the reward and terminal functions for each.
Suppose that the $L_1$-agent is in state $\tilde{s}_\text{init} \in \tilde{\mathcal{S}}$ and takes action $\tilde{a} \in \tilde{\mathcal{A}}$. Further suppose that $\tilde{s}_\text{goal} \in \tilde{\mathcal{S}}$ is the intended result of applying action $\tilde{a}$  to state $\tilde{s}_\text{init}$.
This high-level action causes the execution of an $L_0$-policy with the following modified terminal set and reward function: 
\begin{equation}
\begin{aligned}
\mathcal{E}_\text{episode} = \mathcal{E} \cup \{ s \in \mathcal{S} : F(s) \neq \tilde{s}_\text{init} \} \\
\mathcal{R}_\text{episode}(s, a, s^\prime) = \begin{cases}
1 & \text{ if } F(s^\prime) = \tilde{s}_\text{goal} \\
0 & \text{ else}. 
\end{cases}
\end{aligned}
\end{equation}
Notice that the $L_0$ reward function ignores the ground-environment reward function, $\mathcal{R}$.
This information is instead passed to the $L_1$ reward function.
Denote the rewards accrued over $T$ steps of the $L_0$-episode as $\tilde{r} = \sum_{t=1}^T \mathcal{R}_t$, denote whether the $L_0$-environment terminated as $\tilde{e}$, and denote the final $L_0$-state as $s_\text{term}$.
At the termination of the $L_0$-episode, these quantities are returned to the $L_1$-agent to provide a complete experience tuple $\langle \tilde{s}_\text{init}, \tilde{a}, \tilde{r}, F(s_\text{term}), \tilde{e}\rangle$. 

\section{Learning}

In the previous sections, we defined the semantics of our AMDP hierarchy but did not specify the precise learning algorithms to be used for the $L_1$ and $L_0$-agents.
Indeed, any reinforcement learning algorithm could be used for either of these agents since each operates on a classical MDP.
In our work, we chose to use a deep reinforcement learning method for the $L_0$ learner to process the high-dimensional pixel input and a model-based algorithm for the $L_1$ learner to exploit its long-term planning capabilities.

\subsection{Low Level Learner} \label{section:l0}
As described above, every transition between two $L_1$ states is represented by an $L_0$ AMDP.
So, if there are multiple hundred $L_1$ states and each one has a few neighboring states, there could be hundreds or thousands of $L_0$ AMDPs.
Each $L_0$ AMDP could be solved using a vanilla DQN, but it would take millions of observations to train each one  to learn since every DQN would have to learn from scratch.
To avoid this high computational cost, we share all parameters, except for those in the last fully connected layer of our network, between policies. For each policy we use a different set of parameters for the final fully connected layer. This encourages sharing high-level visual features between policies and imposes that the behavior of an individual $L_0$-policy is  specified by these interchangeable, final-layer parameters.
In our implementation, we used the Double DQN loss \citep{van2016deep} with the Mixed Monte-Carlo update as it has been shown to improve performance on sparse-reward domains \citep{ostrovski2017count}.

Because we share all layers of the network between the DQNs, updating one network could change the output for another.
This can sometimes lead to forgetting policies.
To correct for this, we use an $\epsilon$-greedy policy where we dynamically change epsilon based on how successful the $L_0$ AMDP is.
We measure the success of each $L_0$ AMDP by periodically evaluating them (by setting $\epsilon = 0.01$) and measuring the number of times the policy terminates at the goal state, $\tilde{s}_{goal}$.
We then set $\epsilon$ equal to 1 minus the proportion of the time the $L_0$ AMDP succeeds when evaluated (with a minimum epsilon of $0.01$).
We found this allows the agent to keep exploring actions that were not yet learned or have been forgotten, while exploiting actions that have already been learned.
However, when the transition cannot be consistently completed by a random policy, this method tends to fail.

\subsection{High Level Learner}
For our $L_1$-agent, we use a tabular R-Max learning agent \citep{brafman2002r}.
We chose this reinforcement learning algorithm for our $L_1$-agent as it constructs long-term plans to navigate to under-explored states.
Particularly, every action $\tilde{a} \in \tilde{A}$ is given an R-Max reward until that action has been tried some number of times.
We chose $100$ for this number to ensure that a random policy could discover all possible next abstract states.


It is possible for $L_1$ actions to continue running forever if the agent never transitions between $L_1$ states.
Thus, in practice we only run an $L_1$ action for a maximum of 500 steps.

\begin{algorithm}[t]
    \caption{Object-Oriented AMDP algorithm}
    \label{OOAMDP}
    \begin{algorithmic}[1]
        \Procedure{Learn}{}
        	\State $\mathcal{S}, \mathcal{A} \gets \emptyset$
            \While{training}
                \State $s \gets $ current environment state
                \If{$s \not \in \mathcal{S}$}
                	\State \texttt{Add\_State}$(s)$
                \EndIf
                \State $a \gets \argmax_{a}(Q(s,a)) $
                \State $s', r, t \gets $ perform action $a$
                \State $d_{result} \gets \texttt{Diff}(s, s')$
                \If{$(d_{result}, p_1(s^\prime), \ldots, p_L(s^\prime)) \not \in \mathcal{A}$}
                	\State \texttt{Add\_Action}$(d_{result}, p_1(s^\prime), \ldots, p_L(s^\prime))$
                \EndIf
                \State add $\langle s, a, s', r, t \rangle$ to transition table
                \State run \texttt{Value\_Iteration}
            \EndWhile
        \EndProcedure
        \Procedure{Value\_Iteration}{}
            \For{Some number of steps}
            	\For{$s \in \mathcal{S}$}
                	\For{$a \in $ all applicable actions for $s$}
                    	\State $s' \gets $ apply \texttt{Diff} of $a$ to $s$
                    	\State $Q_t(s, a) \gets \sum_{d_j \in N(a)} \mathcal{T}(a, d_j) [\mathcal{R}(a, d_j) + \gamma V_{t-1}(s)(1 - \mathcal{E}(a, d_j))]$ \Comment{Bellman update}
                    \EndFor
           		\State $V_t(s) \gets \max_{a}(Q_t(s, a))$
                \EndFor
            \EndFor
        \EndProcedure
    \end{algorithmic}
\end{algorithm}

\subsection{Exploration for $L_1$ and $L_0$ Agents}

In this work, we assume the agent is given only the state projection function, $F$, minimizing the work the designer needs to do.
However, this means that the agent must learn the transition dynamics of the $L_1$ AMDP and build up the hierarchy on-the-fly.

To do so, our agent begins with an empty set of states and actions, $\tilde{\mathcal{S}}$ and $\tilde{\mathcal{A}}$.
Because we do not know the transition graph, every state needs to be sufficiently explored in order to find all neighbors.
To aid in exploration, we give every state an \textit{explore} action, which is simply an $L_0$ AMDP with no goal state.
Whenever a new state-state transition is discovered from $\tilde{s}_{1}$ to $\tilde{s}_{2}$, we add a new $L_1$ AMDP action with the initial state $\tilde{s}_{1}$ and goal state $\tilde{s}_{2}$ to $\tilde{\mathcal{A}}$.
In practice, we limit each explore action to being executed $N_\text{explore}$ times.
After being executed $N_\text{explore}$ times, we remove that explore action, assuming that it has been sufficiently explored.
We use $N_\text{explore}=100$ in our experiments. 
The pseudo code is detailed in Algorithm \ref{OOAMDP}.

\section{Constructing an Abstraction}
\label{sec:constructing_an_abstraction}

\begin{figure}
  \centering
  \captionsetup[subfigure]{labelformat=empty}
  \begin{subfigure}[b]{0.35\textwidth}
    \includegraphics[width=\textwidth]{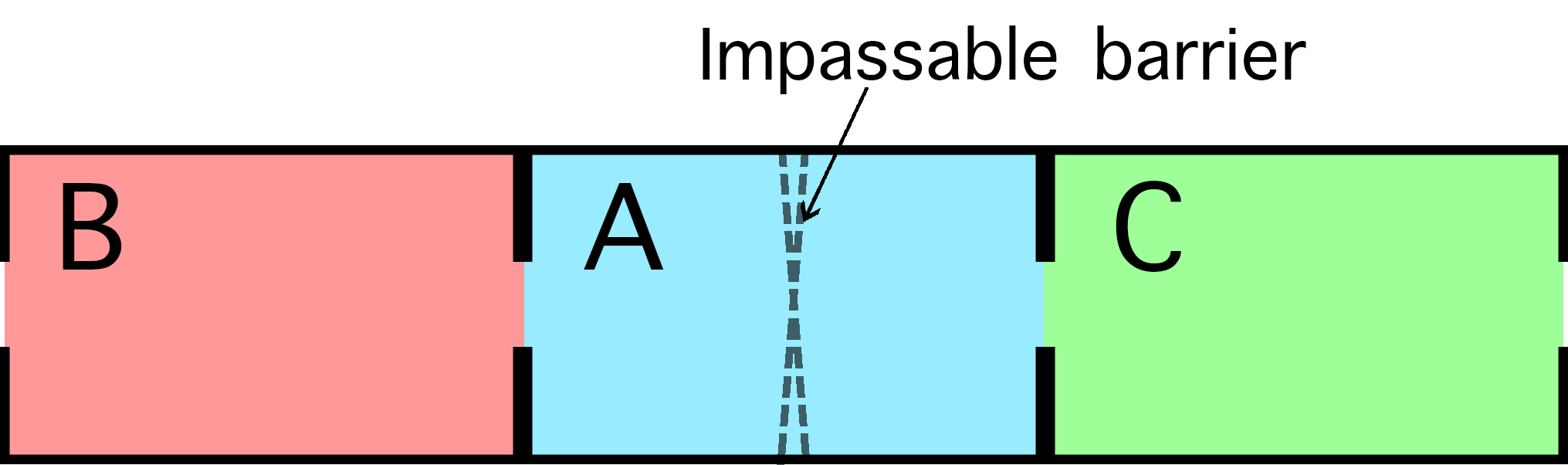}
  \end{subfigure}
  \caption{ Example of a non-Markovian abstraction. The transition dynamics of room A depend on the side from which the agent enters the room.}
  \label{fig:rooms_with_barrier}
\end{figure}

The main benefit of our abstractions is to shorten the reward horizon of the low-level learner.
The guiding principal is to construct an abstraction such that  $L_1$-states encompass small collections of $L_0$-states.
This ensures that the $L_0$-agents can reasonably experience rewards from transitioning to all neighboring $L_1$-states.
It is crucial that the abstraction is as close to Markovian as possible: the transition dynamics for a state should not depend on the history of previous states.
For example, imagine a four rooms domain where room A connects to rooms B and C (Figure \ref{fig:rooms_with_barrier}).
If for some reason there is an impassable wall in room A, then the agent can transition from A to B on one side of the wall and from A to C on the other side.
So depending on how the agent entered the room (the history), the transition dynamics of room A would change.
However, since the high-level learner has seen the agent transition from room B to A and A to C, it would think B and C are connected through A.
The solution would be to divide room A into two smaller rooms split by the impassable barrier.

In our experiments, we split rooms up into smaller sectors in the abstraction to decrease the horizon for the $L_0$ learners and, in some games, to retain the Markovian property of the abstraction.
For Toy MR, these sectors were hand-made for each of the rooms (Figure \ref{fig:toy_mr_map:map_sectors}).
We constructed the sectors such that there were more sectors on the ``tight-ropes,'' areas that required many correct actions to traverse and a single incorrect action would result in a terminal state.
For the Atari experiments, we made square $n \times n$ grids of each of the rooms based on the coordinates of the agent: if the agent is in the top left corner of the screen, it is in sector $1$. If it is in the bottom-right corner, sector $n^2$ (Figure \ref{fig:atari_screens}).
For MR, we chose the grid to be $3 \times 3$.
For Venture, we chose the grid to be $3 \times 3$ inside each of the rooms and $4 \times 4$ in the hallway, as the state-space in the hallway is much larger.
We chose this particular gridding because it is both simple to implement and approximately Markovian across the game's different rooms.
Note that any sufficiently fine-grained sector scheme would perform equivalently.
Accordingly, our particular choice of sector scheme should be regarded as arbitrary.
Other abstractions could be used as long as they are also approximately Markovian.

\section{Experiments}

The aim of our experiments was to assess the effectiveness of our algorithm on complex domains that involve long horizons, sparse rewards, and high-dimensional inputs.
We trained our agents for $50$ million frames.
As in \citet{mnih2015human}, every one million frames, we evaluated our agents for a half a million frames, recording the average episode reward over those evaluation frames. The source code of our implementation is available online\footnote{Code: github.com/chrisgrimm/deep\_abstract\_q\_network}. 

\subsection{Baselines}
We chose two baselines to compare against our algorithm: Double DQN \citep{van2016deep} and Pseudo-Count based IM \citep{Bellemare2016UnifyingCE}, both using the Mixed Monte-Carlo return \citep{ostrovski2017count}.
We chose Double DQN as it performed very well on many Atari games, but has not been optimized for exploration.
We chose the IM agent as it explored the highest the number of rooms in Montezuma's Revenge to the best of our knowledge.
One of the key aspects to the success of this algorithm, that was not required for our algorithm, was giving the agent multiple \textit{lives}, which was discussed in our Related Work section.
We, therefore, also compared to the IM agent with this addition.

We tested our algorithm against these baselines in three different domains.
It is important to note that we do provide the factorized state projection function and the set of predicate functions.
However, in many real world domains, there are natural decompositions of the low-level state into abstract components, such as the current room of the agent in the room navigation task. 

For the toy domains and Single-Life MR (described below) we used our own implementation of pseudo-counts \citep{Bellemare2016UnifyingCE} as the authors were unwilling to provide their source code.
Our implementation was not able to perform at the level of the results reported by \citeauthor{Bellemare2016UnifyingCE}, only discovering 7-10 rooms on Atari Montezuma's Revenge in the time their implementation discovered 15 (50 million frames).
Our implementation still explores more rooms than our baseline, Double DQN, which only discovered 2 rooms.
We contacted other researchers who attempted to replicate these results, and they were likewise unable to.
\citeauthor{Bellemare2016UnifyingCE}, however, did kindly provide us with their raw results for Montezuma's Revenge and Venture.
We compared against these results, which were averaged over 5 trials.
Due to our limited computing resources, our experiments were run for a single trial.

\subsection{ Four Rooms and Toy Montezuma's Revenge }

\begin{figure}[htb!]
  \centering
  \begin{subfigure}[b]{0.15\textwidth}
    \vspace{-3mm}
    \includegraphics[width=\textwidth]{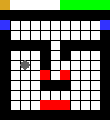}
    \caption{Example Screen}
    \label{fig:toy_mr_map:example_state}
  \end{subfigure}
  \hspace{10mm}
  \begin{subfigure}[b]{0.17\textwidth}
    \includegraphics[width=\textwidth]{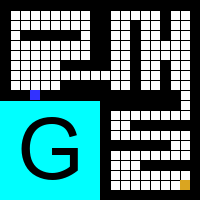}
    \caption{Map of Four Rooms}
    \label{fig:toy_mr_map:four_map}
  \end{subfigure}
  \vspace{4mm}
  \begin{subfigure}[b]{0.45\textwidth}
    \includegraphics[width=\textwidth]{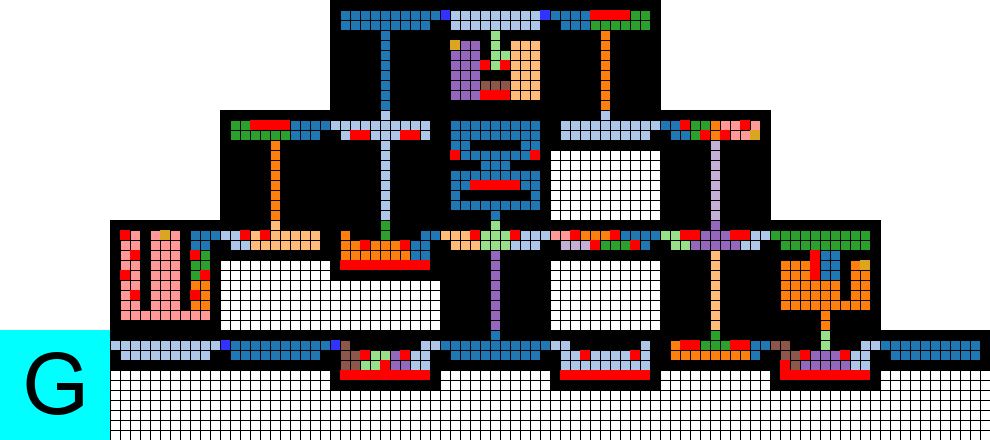}
    \caption{Map of all rooms in Toy MR with color-coded sectors}
    \label{fig:toy_mr_map:map_sectors}
  \end{subfigure}
  \caption{\ref{fig:toy_mr_map:example_state} Example screen that is common across Four Rooms and Toy MR. The yellow square at the top left represents that the agent is holding a key and the green bar on the right represents the agent's remaining lives.
  \ref{fig:toy_mr_map:four_map}, \ref{fig:toy_mr_map:map_sectors} The map of all the rooms in Four Rooms and Toy MR. Blue squares are locked doors, yellow squares are keys that can unlock the doors, and the red squares are traps that result in a terminal state (or the loss of a life when playing with lives). The teal room with the `G' is the goal room. Entering this room gives the agent a reward of 1 (the only reward in the game) and results in a terminal state. The sectors provided to the agent in Toy MR are color-coded.
  } 
  \label{fig:toy_mr_map}
\end{figure}

We constructed a toy version of the room navigation task: given a series of rooms, some locked by doors, navigate through the rooms to find the keys to unlock the doors and reach the goal room.
In this domain, each room has a discrete grid layout.
The rooms consist of keys (gold squares), doors (blue squares), impassible walls (black squares), and traps that end the episode if the agent runs into them (red squares).
The state given to the agent is the pixel screen of the current room, rescaled to 84x84 and converted to gray-scale.
We constructed two maps of rooms: \textit{Four Rooms} and \textit{Toy Montezuma's Revenge} (Toy MR).
Four Rooms consists of three maze-like rooms and one goal room (Figure \ref{fig:toy_mr_map:four_map}).
Toy MR consists of $24$ rooms designed to parallel the layout of the Atari Montezuma's Revenge (Figure \ref{fig:toy_mr_map:map_sectors}).
In the Four Rooms domain, the game terminates after $10\,000$ steps, while in Toy MR, there is no limit on the number of steps.

The abstraction provided to the agent consists of 10 attributes: the location of the agent, a Boolean for the state of each key (4 keys total) and each door (4 doors total), and the number of keys the agent had.
The location of the agent consists of the current room and sector.
We used sectors for Toy MR to decrease the horizon for each $L_0$ learner (as detailed in the Section \ref{sec:constructing_an_abstraction}), but not for Four Rooms since it does not have deadly traps that hinder exploration.
Although the sectors seem to divide much of the state-space, the low-level learners remain crucial to learning the policies to navigate around traps and transition between high-level states.

\begin{figure}[htb!]
  \centering
  \begin{subfigure}[b]{0.2\textwidth}
    \includegraphics[width=\textwidth]{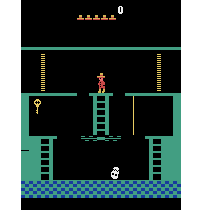}
    \caption{MR}
  	\label{fig:atari_screens:mr}
  \end{subfigure}
  \begin{subfigure}[b]{0.2\textwidth}
    \includegraphics[width=\textwidth]{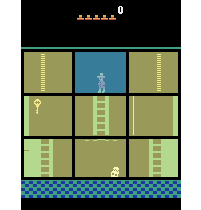}
    \caption{MR Sectors}
  	\label{fig:atari_screens:mr_sectors}
  \end{subfigure}
  \begin{subfigure}[b]{0.2\textwidth}
    \includegraphics[width=\textwidth]{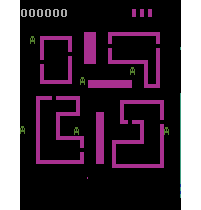}
    \caption{Venture}
  	\label{fig:atari_screens:venture}
  \end{subfigure}
  \begin{subfigure}[b]{0.2\textwidth}
    \includegraphics[width=\textwidth]{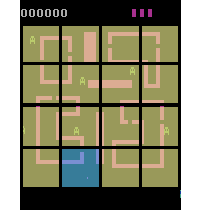}
    \caption{Venture Sectors}
  	\label{fig:atari_screens:venture_sectors}
  \end{subfigure}
  \caption{ \ref{fig:atari_screens:mr}, \ref{fig:atari_screens:venture} Example screens of Atari 2600 Montezuma's Revenge (MR) and Venture.
  \ref{fig:atari_screens:mr_sectors}, \ref{fig:atari_screens:venture_sectors} Illustrations of the sectors we constructed for both a room in MR and the hallway in Venture.
  The sector the agent is currently occupying is in blue, the other possible sectors are in yellow.}
  \label{fig:atari_screens}
\end{figure}

\begin{figure}
  \centering
  \captionsetup[subfigure]{labelformat=empty}
  \begin{subfigure}[b]{0.45\textwidth}
    \includegraphics[width=\textwidth]{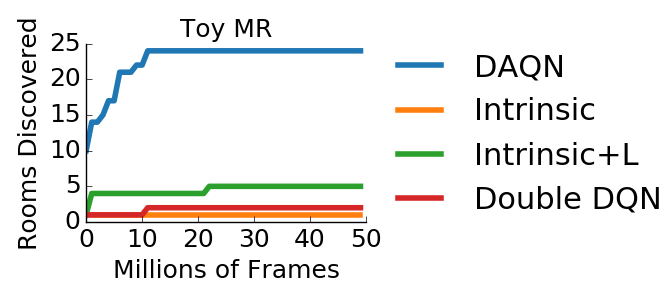}
  \end{subfigure}
  \caption{ Rooms discovered in the Toy MR domain using the Double DQN, DAQN, IM, and IM with a 5-lives variant of Toy MR (Intrinsic+L).}
  \label{fig:rooms_results}
\end{figure}

\begin{figure*}
  \centering
  \includegraphics[width=\textwidth]{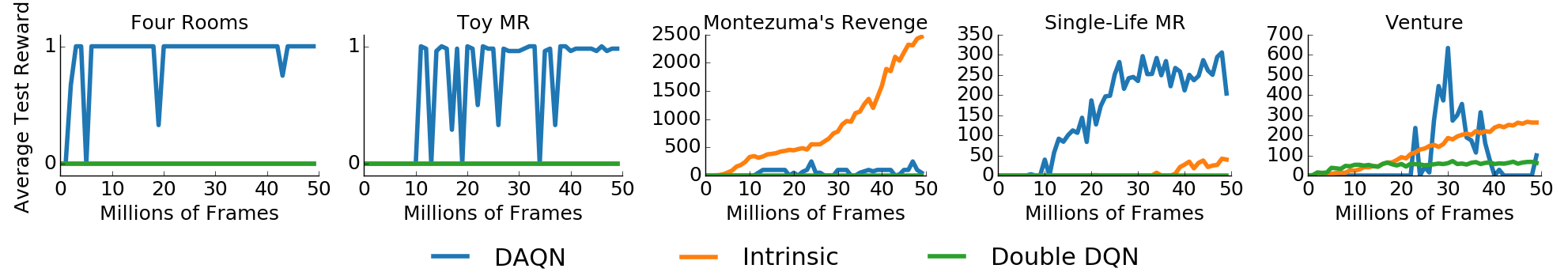}
  \caption{ Average reward in the Four Rooms, Toy MR, Atari MR, Single-Life Atari MR, and Atari Venture domains using the following models: DAQN (blue), Double DQN (green) and IM (orange).
  In Four Rooms and Toy MR, both IM and Double DQN fail to score an average reward above zero, and are thus overlapping.
  We use the raw IM and Double DQN data from \citet{Bellemare2016UnifyingCE} on Montezuma's Revenge and Venture.
  All other plots show our implementations' results.}
  \label{fig:reward_results}
\end{figure*}

Our results (Four Rooms and Toy MR plots in Figure \ref{fig:reward_results}) show that for both domains, Double DQN and the IM agent failed to learn to complete the game, while our agent learned to consistently solve both toy problems.
On the Toy MR domain, both agents fail to escape the first room when the agent is only provided one life.
This reflects the issue with pseudo-counts for IM that we described previously: that the image is factored in a way that makes the key and agent pixels independent, with the result that the exploration bonuses of backtracking to the doors are lower than those of remaining near the key.
In contrast, our agent was not only able to explore all the rooms in Toy MR, but also to learn the complex task of collecting the key to unlock the first room, collecting two more keys from different rooms and then navigating to unlock the final two doors to the goal room (Figure \ref{fig:rooms_results}). 

We emphasize that this marked difference in performance is due to the different ways in which each method explores.
Particularly, our DAQN technique is model-based at the high-level, allowing our coupled agents to quickly generate new long-term plans and execute them at the low-level.
This is in contrast to IM, which must readjust large portions of the network's parameters in order to change long-term exploration policies. 

\subsection{Montezuma's Revenge Atari 2600}

Montezuma's Revenge (MR) is an Atari game very similar to the rooms and doors toy problems: there is a series of rooms, some blocked by doors, and keys are spread throughout the game.
There are also monsters to avoid, coins that give points, and time-based traps, such as bridges over lava pits that disappear and reappear on a timer.

Our abstraction had a similar state-space to Toy MR, consisting of 12 attributes: the location of the agent, a Boolean attribute for the presence of each key (4 keys total) and each door (6 doors total), and the number of keys.
The location of the agent consists of the current room and sector.
We created coarse sectors based on the agent's location in a room by gridding each room into nine equal square regions.
We prevented sector transitions while the agent was falling to avoid entering a sector and immediately dying from falling.
As an example, consider the agent in Figure \ref{fig:atari_screens:mr}.
Figure \ref{fig:atari_screens:mr_sectors} illustrates the sector that the agent occupies.
The abstraction of this state would be: Room $1$ (the starting room) and Sector $(1, 2)$ with no keys collected or doors unlocked.

We also tested the DAQN on MR where the agent is only given a single life (i.e. the environment terminates after a single death).
Normally in MR, when the agent dies, it returns to the location from which it entered the room (or the starting location in the first room) and retains the keys it has collected.
Because of this, a valid policy for escaping the first room is to navigate to the key, collect it, and then purposefully end the life of the agent.
This allows the agent to return to the starting location with the key and easily navigate to the adjacent doors.
In this single life variant, the agent cannot exploit this game mechanic and, after collecting the key, must backtrack all the way to the starting location to unlock one of the doors.
This comparison illustrates our algorithm's ability to learn to separate policies for different tasks.

With lives, our algorithm did not discover as many rooms as the IM agent since our agent was not able to traverse the timing-based traps.
These traps could not be traversed by random exploration, so our agent never learned that there is anything beyond these traps.
Our agent discovered six rooms out of the total 24 -- all the rooms that can be visited without passing these traps.

Our agent underperformed in Atari Montezuma's Revenge (Montezuma's Revenge plot in Figure \ref{fig:reward_results}) because of timing based traps that could not be easily represented in a discrete high-level state space.
However, when we grant our agent only one life, our method greatly outperforms previous methods: not only was our agent able to escape the first room, but it also discovered five more, while the Double DQN and IM agents are not able to escape the first room (Single-Life MR plot in Figure \ref{fig:reward_results}).
This is because the one-life setting necessitates backtracking-like behavior in a successful policy.
As we mentioned before, the IM agent is incapable of learning to backtrack and thus cannot perform in this setting.
We emphasize that this inability arises on account of the pseudo-count probabilistic model treating the location of the agent and the presence of the key as independent.
This property actively discourages the agent from backtracking because backtracking would lead to states with higher pseudo-counts and, thus, less intrinsic reward.


\subsection{Venture Atari 2600}
Venture is a game that consists of four rooms and a hallway. 
Every room contains one item.
The agent must navigate through the hallway and the rooms, avoiding monsters, to collect these items.
Once an item is collected and the agent leaves the room, that room becomes locked.

Our abstraction for this game consisted of 9 attributes: the location of the agent, a Boolean \textit{locked} attribute for each room (4 rooms total), and a Boolean for whether the item in the current room has been collected (4 items total).
The location of the agent consists of the current room and sector.
Sectors were constructed with a coarse $3 \times 3$ gridding of each room and a $4 \times 4$ gridding of the hallway.
As an example, in Figure \ref{fig:atari_screens:venture} the agent is the the small pink dot at the bottom of the screen and Figure \ref{fig:atari_screens:venture_sectors} shows the sector the agent occupies.
In this state, the abstraction would be: Room $8$ (the hallway) and Sector $(1, 0)$ with no items collected.

In this experiment, we receive a much higher evaluation performance than both of our baselines (Venture plot in Figure~\ref{fig:reward_results}), illustrating our agents ability to execute and learn long-term plans.
At around 30 million frames, our agent's performance greatly decreases.
This performance drop is due to our agent exploring further into new rooms and training the sub-policies to reach those new rooms.
Since the sub-policies for exploitation are not trained during this time, as the DQN weights higher up in the network are updated to train the exploration sub-policies, the exploitation sub-policies are forgotten.
Once the agent finishes exploring all $L_1$ states, we would expect the agent would revisit those exploitation sub-policies and relearn them.

\section{Discussion and Future Work}


In this paper, we presented a novel way of combining deep reinforcement learning with tabular reinforcement learning using DAQN.
The DAQN framework generally allows our agent to explore much farther than previous methods on domains and exploit robust long-term policies.

In our experiments, we showed that our DAQN agent explores farther in most high-dimensional domains with long-horizons and sparse reward than competing approaches.
This illustrates its capacity to learn and execute long-term plans in such domains, succeeding where these other approaches fail.
Specifically, the DAQN was able to learn backtracking behavior, characteristic of long-term exploration, which is largely absent from existing state-of-the-art methods.

The main drawback to our approach is the requirement for a hand-annotated state-projection function that nicely divides the state-space.
However, for our method allows this function need only specify abstract states, rather than abstract transitions or policies, and thus requiring minimal engineering on the part of the experimenter.
In future work, we hope to learn this state-projection function as well.
We are exploring methods to learn from human demonstration, as well as methods that learn only from a high-level reward function.
Ultimately, we seek to create compositional agents that can learn layers of knowledge from experience to create new, more complex skills.
We also plan to incorporate a motivated exploration algorithm, such as IM \citep{Bellemare2016UnifyingCE}, with our $L_0$ learner to address our difficulty with time-based traps in MR.

Our approach also has the ability to expand the hierarchy to multiple levels of abstraction, allowing for additional agents to learn even more abstract high-level plans.
In the problems we investigated in this work, a single level of abstraction was sufficient, allowing our agent to reason at the level of rooms and sectors.
However, in longer horizon domains, such as inter-building navigation and many real-world robotics tasks, additional levels of abstraction would greatly decrease the horizon of the $L_1$ learner and thus facilitate more efficient learning. 



\begin{acks}
  This material is based upon work supported by the National Science Foundation under grant numbers IIS-1426452, IIS-1652561, and IIS-1637614, DARPA under grant numbers W911NF-10-2-0016 and D15AP00102, and National Aeronautics and Space Administration under grant number NNX16AR61G.
\end{acks}


\bibliographystyle{ACM-Reference-Format}  
\bibliography{ref}  

\end{document}